\pdfoutput=1
\documentclass[11pt]{article}

\usepackage[final]{acl}

\usepackage{booktabs}

\usepackage{times}
\usepackage{latexsym}

\usepackage[T1]{fontenc}
\usepackage[utf8]{inputenc}

\usepackage{microtype}

\usepackage{inconsolata}

\usepackage{graphicx}
\usepackage{enumerate}  
\title{{Can Post-Training Quantization Benefit from an Additional QLoRA Integration?}}

\author{Xiliang Zhu$^*$,\ Elena Khasanova$^*$, \ Cheng Chen$^*$ \\
          Dialpad Inc. \\
  \texttt{\{xzhu,elena.khasanova,cchen\}@dialpad.com}}

\begin{document}
\maketitle 
\begin{abstract}
Large language models (LLMs) have transformed natural language processing but pose significant challenges for real-world deployment. These models necessitate considerable computing resources, which can be costly and frequently unavailable. Model compression techniques such as quantization are often leveraged to alleviate resource demand, but they may have a negative impact on the generation quality. In this study, we explore the integration of 4-bit Post-training Quantization (PTQ) with QLoRA \cite{DBLP:conf/nips/DettmersPHZ23} to address these issues. We demonstrate through extensive experiments that this integration outperforms standard PTQ, and in some cases even 16-bit full-parameter fine-tuning on LLMs, validated across proprietary and public datasets with different quantization algorithms. The results demonstrate the efficacy of PTQ-QLoRA integration, offering a viable solution for deploying powerful LLMs in resource-constrained environments without compromising on performance.
\end{abstract}

\def\thefootnote{*}\footnotetext{Equal Contributions. Sorted by Last Name in reverse order.}\def\thefootnote{\arabic{footnote}}

\section{Introduction}
Large language models (LLMs) have undeniably revolutionized the field of natural language processing and keep growing in both popularity and size. However, the “large” in LLMs is both their benefit and their curse. As the models are becoming more powerful, they are increasingly harder to train, deploy and serve in real-life applications in industry. They require substantial computing resources which are not only expensive but also not always readily available. 

Obtaining resources for training LLMs is a challenge of its own, but deploying LLMs in customer-facing applications poses a new set of challenges. Specifically, LLM inference in real-life scenarios comes with certain challenges. It must meet latency requirements to ensure a smooth user experience for end users. It is also subject to memory constraints from accessible hardware, which is not always optimized for LLMs. Additionally, it needs to allow for frictionless scaling as the number of requests to LLMs grows with the number of users or features it serves. Therefore, there exists a need for optimization techniques that would allow for deployment of the most powerful LLMs regardless of the number of parameters but also address these issues without significant loss in performance. 

One of the popular techniques to optimize memory usage and computational efficiency is quantization, which reduces the precision of the numerical representation of data and thereby the model’s size and the computational resources required for inference by a large margin, but often results in meaningful accuracy loss \cite{10.5555/3618408.3618715}. At the same time, quantized large models can outperform full-precision models of smaller size \cite{lee2024comprehensive}, making quantized models a potentially preferred option and recovering accuracy loss a particularly important task. 

In this study, we explore the integration of Post-training Quantization (PTQ) and QLoRA \cite{DBLP:conf/nips/DettmersPHZ23}, which utilizes parameter-efficient fine-tuning (PEFT) on a quantized model, to mitigate the loss in accuracy due to quantization. We focus solely on 4-bit quantization because it provides an optimal balance of memory footprint, latency and accuracy for our specific use cases, where the model is deployed\footnote{Served by Nvidia T4} to handle business conversations such as support calls or meetings.  We show through experiments that this integration outperforms simple post-training quantization and in certain cases even the 16-bit fully fine-tuned model.

Our contributions are the following:
\begin{itemize}
  \item We explore the integration of 4-bit Post-training Quantization (PTQ) with QLoRA, delivering task performance that matches or surpasses 16-bit full fine-tuning on LLMs.
  \item We examine the proposed integration with extensive experiments involving multiple base LLMs and quantization methods, accompanied by a detailed performance comparison.
  \item To ensure a robust evaluation of this integration, we perform experiments using:
  \begin{enumerate}[(i)]
      \item a proprietary dataset with real-world Automatic Speech Recognition (ASR)-generated transcription data from real-world business conversations
      \item an open dataset from the business domain. We test our approach on both generation and classification tasks.
  \end{enumerate}
\end{itemize}

\begin{figure*}[t]
\centering
\includegraphics[width=1\textwidth, height=5.5cm]{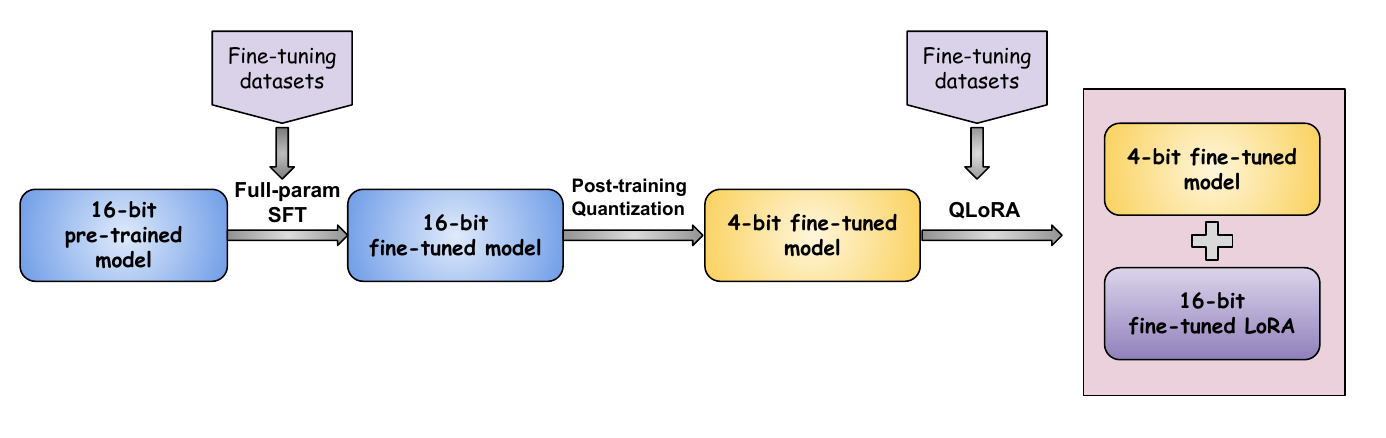}
\caption{\small Diagram of the PTQ-QLoRA integration. Note that we apply the same fine-tuning datasets twice during full-parameter SFT and QLoRA fine-tuning respectively.}
\label{fig1}
\end{figure*}

\section{Background}
Traditionally, deep neural network models utilize high-precision floating point numbers to represent weights and activations, which requires significant memory and computational resources. Quantization has emerged as a powerful technique to address this challenge by quantizing floating-point representations into a lower bit-width, effectively reducing the model's memory footprint and computational cost.

Quantization techniques generally fall into two main categories: Post-training Quantization (PTQ) and Quantization-aware Training (QAT). The former quantizes a model after the training is complete, without the need for retraining. Early work like \cite{Jacob_2018_CVPR} proposed a quantization schema that uses integer arithmetic to approximate the floating-point. \cite{10.5555/3524938.3525605} computes a layer-wise local loss and optimizes this loss with a soft relaxation. \cite{Li2021BRECQPT} proposed BRECQ framework which achieves a good balance between cross-layer dependency and generalization error by reconstructing at the block granularity. More recently, LLM.int8() from \cite{10.5555/3600270.3602468} demonstrated for the first time that multi-billion parameter transformers can be effectively quantized to Int8. Moreover, \cite{frantar-gptq} introduced GPTQ which can accurately quantize LLMs of billions of parameters to 3-4 bits per component. Activation-aware Weight Quantization (AWQ) from \cite{MLSYS2024_42a452cb} employs per-channel scaling to reduce the quantization loss of salient weights.

Conversely, QAT techniques typically involve retraining the model with quantized parameters so that the model can converge to a point with better loss \cite{Gholami2021ASO}. \cite{Nagel2021AWP} presented a standard QAT pipeline that leads to near-floating-point accuracy results for a wide range of models.

Another efficient approach to adapting pre-trained models with minimal overhead is Parameter-Efficient Fine-tuning (PEFT). One direction is adapter-based method, which injects small adapter modules into pre-trained models \cite{pfeiffer-etal-2020-mad}\cite{Houlsby2019ParameterEfficientTL}. More recently, Low-Rank Adaptation (LoRA) \cite{hu2022lora} has become increasingly popular, which greatly reduces the number of trainable parameters by introducing rank decomposition matrices. Moreover, QLoRA \cite{DBLP:conf/nips/DettmersPHZ23} backpropagates gradients through a quantized model into LoRA while preserving high task performance. Although \cite{DBLP:conf/nips/DettmersPHZ23} shows QLoRA can match the accuracy of 16-bit full fine-tuning in T5 \cite{raffel2023exploringlimitstransferlearning} and RoBERTa \cite{liu2019robertarobustlyoptimizedbert}, the comparison of QLoRA and 16-bit tuning on other larger language models has not been studied to our best knowledge.

\section{Methodology}
\subsection{Overview}
\label{sec:method_overview}
Figure \ref{fig1} illustrates the PTQ-QLoRA integration. Our steps are as follows:

\begin{enumerate}
    \item We first employ full-parameter supervised fine-tuning (SFT) using a mixture of general instruction-following data and our internal tasks' training data on a pre-trained model, to obtain the fine-tuned model (in 16-bit).
    \item We then apply 4-bit Post-training Quantization (PTQ) on the 16-bit fine-tuned model, to obtain the quantized 4-bit model.
    \item Lastly,  we leverage the QLoRA~\cite{DBLP:conf/nips/DettmersPHZ23} approach to do another round of SFT on the quantized 4-bit model through a LoRA~\cite{DBLP:conf/iclr/HuSWALWWC22}.
\end{enumerate}

\subsection{Models}
\label{sec:method_models}
In this study, we employ three commonly-adopted pre-trained open models:
\begin{itemize}
  \item \textbf{LLaMA2-7B\footnote{\url{https://huggingface.co/meta-llama/Llama-2-7b-hf}, accessed August 2024}:} The LLaMA2 series of LLM models \cite{touvron2023llama2}  developed by Meta.
  \item \textbf{Qwen2-7B}\footnote{\url{https://huggingface.co/Qwen/Qwen2-7B}, accessed August 2024}. The Qwen2 series LLMs \cite{bai2023qwen,yang2024qwen2technicalreport} from Alibaba, supporting long context lengths with strong performance on various %Chinese and English 
benchmarks.
  \item \textbf{Mistral-7b-v0.3}\footnote{\url{https://huggingface.co/mistralai/Mistral-7B-v0.3}, accessed August 2024}. The Mistral series models \cite{jiang2023mistral} are proposed by Mistral AI. It leverages grouped-query and sliding window attention to effectively handle long sequences.
\end{itemize}

Pre-trained base versions of the three models are selected for our experiments rather than their instruction-tuned variations for several reasons. Firstly, it is often easier to ``steer" the behavior of the base models using limited in-domain training data, and our internal findings indicate that when fine-tuned for our internal downstream tasks, the base models consistently demonstrate superior performance (about $5\%$ better across all tasks). Secondly, instruction-tuned variants often have extensive preference alignment done on external datasets which may not represent the preference for our use cases. Lastly, specific chat template is often applied to the instruction-tuned variants. We can design our own simplified templates during finetuning the base models to save formatting tokens in inference. Therefore, the detailed comparison of the instruction-tuned variants is out of the scope of this work.

The weights of the models are sourced from HuggingFace~\cite{wolf2019huggingface}. In addition, we opted for the 7B model size due to its ability to strike a favorable balance between performance and latency, especially when deployed in production contexts with 4-bit quantization.

\subsection{Quantization Methods}
\label{sec:quant_methods}
We adopt quantization methods that support fine-tuning LoRA adapters added to a quantized and freezed base model (i.e. QLoRA~\cite{DBLP:conf/nips/DettmersPHZ23}) as of June 2024, which are bitsandbytes (BNB) \footnote{\url{https://github.com/bitsandbytes-foundation/bitsandbytes}} and GPTQ~\cite{DBLP:conf/iclr/FrantarAHA23}. We choose 4-bit quantization for all models. AWQ~\cite{DBLP:conf/mlsys/0002TTYCWXDG024} seems to have a compatibility issue with CUDA environment at the time and thus is not included in our experiments.

\section{Experiment}
\subsection{Datasets}
To demonstrate the effectiveness of the PTQ-QLoRA integration, we perform experiments on both our internal and public benchmarks. While we cannot release the internal datasets nor reveal their details, we provide description on how we curate external datasets, which are publicly available and the results can be reproduced. In addition, as we utilize the pre-trained base model, instruction-following samples from the general domain (\textbf{General Instruction Dataset}) are also incorporated in our fine-tuning processes to ensure the general instruction-following capability of the resulting models. The General Instruction Dataset is produced by the self-instruct methodology \cite{wang-etal-2023-self-instruct} using GPT-4 to obtain diverse task instructions and corresponding responses. More details of our General Instruction Dataset curation process can be found in Appendix \ref{sec:instruct-data}.
\begin{table}
\centering
% \scriptsize
\begin{tabular}{l|c|c|c}
\toprule
Dataset                   & train & dev  & test \\
\midrule
General Instruction      & 50000 & 3000 & N/A  \\
Summarization                 & 6000  & 700  & 700  \\
Action Items                 & 6000  & 700  & 700  \\
Call Purpose                 & 2000  & 300  & 300  \\
Call Outcome                 & 2000  & 300  & 300  \\

DialogSum  & 7000  & 900  & 900  \\
banking77                 & 4500  & 600  & 600  \\
bitext\_customer\_support & 4500  & 600  & 600  \\
\bottomrule
\end{tabular}
\caption{\small Size of the datasets in our experiments.}
\label{tab:external_datasets_stats}
\end{table}

\subsubsection{Internal Task Datasets} 
The internal data source used in this study is real business conversation transcripts generated from our in-house ASR engine. We create four task datasets which include two text generation tasks and two text classification tasks based on our transcription data:

\begin{itemize}
    \item \textbf{Summarization}: Our summarization task is to generate a coherent and concise summary of a given conversation transcript, with varying summary length requirements (long, medium or short) or format (e.g. bullet points) specified in the prompt. 
    \item \textbf{Action Items}: We define our Action Items task as generating a list of unfinished, actionable tasks based on a conversation transcript. Each task is a one-sentence summary of an activity that should occur after the conversation has ended.
    \item \textbf{Call Purpose}: The Call Purpose task aims to classify the conversation's purpose into one of the pre-defined categories.
    \item \textbf{Call Outcome}: The Call Outcome is another classification task that categorizes the outcome of a business conversation into one of the pre-defined categories.
    
\end{itemize}

Details about the prompts used for our internal tasks can be found in Appendix \ref{sec:fine-tuning-prompt}. The labels of our internal task datasets are generated by GPT-4, which are manually reviewed and post-processed to remove samples identified with minor issues. The remaining samples are deemed of high quality overall. 

\subsubsection{External Tasks Datasets} 
Since we cannot reveal our internal datasets, we select a set of public datasets to validate our results and to show that our observations can be reproduced using publicly available datasets:
\begin{itemize}
    \item \textbf{knkarthick/dialogsum\footnote{\url{https://huggingface.co/datasets/knkarthick/dialogsum}, accessed August 2024}:} This dataset~\cite{chen-etal-2021-dialogsum} is a large-scale dialogue summarization dataset, consisting of 13,460 dialogues with corresponding manually labeled summaries and topics. To make it similar to our internal summarization task, we use the long/medium/short prompts for each dialogue and use GPT-4 to generate summaries. We set the number of samples of train/dev/test as 7000/900/900.
    \item \textbf{PolyAI/banking77\footnote{\url{https://huggingface.co/datasets/PolyAI/banking77}, accessed August 2024.}:} This dataset~\cite{Casanueva2020} consists of online banking queries annotated with their corresponding intents. There are 77 fine-grained intents. The original dataset only has train and test sets. We use a randomly sampled $10\%$ of the train split as the development set. We randomly shuffle the intents in the task prompts, and we set the number of samples of train/dev/test as 4500/600/600. These pre-processing steps are done to make it more similar to our internal tasks. 
    \item \textbf{bitext/Bitext-customer-support-llm-chatbot-training-dataset\footnote{ \url{https://huggingface.co/datasets/bitext/Bitext-customer-support-llm-chatbot-training-dataset}, accessed August 2024}:} This hybrid synthetic dataset has 27 intents assigned to 10 categories. The categories and intents have been selected from Bitext's collection of 20 vertical-specific datasets, covering the intents that are common across all 20 verticals. The original dataset only has a train split. We divide it into train/dev/test following 8/1/1 split ratio, and set the number of samples of train/dev/test as 4500/600/600. The intents in the task prompts are also randomly shuffled. Again, these pre-processing steps are done to make it more similar to our internal tasks.
\end{itemize}

\begin{table*}[t!]
\centering
\scriptsize
\resizebox{\linewidth}{!}{%
\begin{tabular}{l|ccccc|cccc|c|c}
\toprule
\multicolumn{1}{c}{} & \multicolumn{5}{c}{\textbf{Summarization}} & \multicolumn{4}{c}{\textbf{Action Items}} & \multicolumn{1}{c}{\textbf{Call Purpose}} & \multicolumn{1}{c}{\textbf{Call Outcome}} \\ 

\cmidrule(r){1-1} 
\cmidrule(r){2-6} \cmidrule(r){7-10} \cmidrule(r){11-11} \cmidrule(r){12-12}

\textbf{Models} & \textbf{R1} & \textbf{R2} & \textbf{RL} & \textbf{RLsum} & \textbf{AlScore} & \textbf{R1} & \textbf{R2} & \textbf{RL} & \textbf{RLsum} & \textbf{F1-micro}     & \textbf{F1-micro}     \\
Qwen2-7b + \textbf{SFT-16bit}   & 0.5534 & 0.2798 & 0.392  & 0.42 & 0.883 & 0.5428 & 0.3408 & 0.4156 & 0.5081 & 0.5953  & 0.7984       \\
Qwen2-7b + \textbf{PTQ-BNB-4bit} & 0.5534 & 0.2774 & 0.3919 & 0.4194 & 0.886 & 0.5387   & 0.3371 & 0.4151 & 0.5061   & 0.6031       & 0.7963       \\
Qwen2-7b + PTQ-BNB-4bit + \textbf{QLoRA}  & \textbf{0.5701}  & \textbf{0.2925} & \textbf{0.4103} & \textbf{0.4352}  & \textbf{0.89} & \textbf{0.5469}  & \textbf{0.3548} & \textbf{0.427}  & \textbf{0.5128}  & \textbf{0.6381}     & \textbf{0.835}   \\
Qwen2-7b + \textbf{PTQ-GPTQ-4bit}  & 0.5493   & 0.2659 & 0.3831 & 0.4081 & 0.887 & 0.5404   & 0.3397 & 0.4199 & 0.5084  & 0.5875   & 0.8004  \\
Qwen2-7b + PTQ-GPTQ-4bit + \textbf{QLoRA} & 0.5654 & 0.2865 & 0.4034 & 0.4271 & 0.888 & 0.5322  & 0.335  & 0.4097 & 0.4984 & 0.6304       & 0.835  \\
\midrule
Llama2-7b + \textbf{SFT-16bit}  & \textbf{0.5755}  & \textbf{0.3038} & \textbf{0.421}  & \textbf{0.4465}  & \textbf{0.889}  & 0.541  & 0.3567 & 0.4205 & 0.5121  & 0.6848       & 0.8554      \\
Llama2-7b + \textbf{PTQ-BNB-4bit} & 0.5597 & 0.2885 & 0.4091 & 0.4352 & 0.887 & 0.5175 & 0.3411 & 0.4023 & 0.4855  & 0.6537 & 0.8554 \\
Llama2-7b + PTQ-BNB-4bit + \textbf{QLoRA} & 0.5695 & 0.2936 & 0.4098 & 0.4349 & 0.875 & 0.5395 & 0.3435 & 0.4103 & 0.5057  & 0.6887       & \textbf{0.8697}       \\
Llama2-7b + \textbf{PTQ-GPTQ-4bit} & 0.5716 & 0.2973 & 0.4136 & 0.4393 & 0.883 & 0.5507 & 0.3631 & 0.4281 & 0.5202 & \textbf{0.6926}       & 0.8554       \\
Llama2-7b + PTQ-GPTQ-4bit + \textbf{QLoRA} & 0.5727 & 0.2978 & 0.4129 & 0.4398 & 0.885 & \textbf{0.5638} & \textbf{0.366}  & \textbf{0.4299} & \textbf{0.5308} & \textbf{0.6926}       & 0.8493       \\
\midrule
Mistral-7b + \textbf{SFT-16bit}   & 0.5738  & 0.3056 & 0.418  & 0.4423 & 0.894 & 0.5459  & 0.34  & 0.4154 & 0.513 & 0.6576       & 0.831        \\
Mistral-7b + \textbf{PTQ-BNB-4bit}  & 0.572 & 0.2998 & 0.4128 & 0.4393 & 0.889 & 0.5367 & 0.3423 & 0.4157 & 0.5064  & 0.7198  & \textbf{0.8635}  \\
Mistral-7b + PTQ-BNB-4bit + \textbf{QLoRA}  & 0.5758  & 0.3075 & \textbf{0.4242} & 0.4466 & 0.891 & 0.5373  & 0.3432 & 0.4118 & 0.5068   & \textbf{0.7237}       & 0.8554  \\
Mistral-7b + \textbf{PTQ-GPTQ-4bit}  & 0.5772 & 0.3057 & 0.4175 & 0.4427 & \textbf{0.895} & 0.4196 & 0.2808 & 0.327  & 0.3967 & 0.6576   & 0.833  \\
Mistral-7b + PTQ-GPTQ-4bit + \textbf{QLoRA} & \textbf{0.5821}  & \textbf{0.3114} & 0.4217 & \textbf{0.4495} & 0.891 & \textbf{0.5465}  & \textbf{0.3554} & \textbf{0.4267} & \textbf{0.5153}    & 0.7082 & 0.8534   \\
\bottomrule
\end{tabular}
}
\caption{\small Performance of different models on our internal task benchmark. R1, R2, RL and RLsum refer to ROUGE-1, ROUGE-2, ROUGE-L and ROUGE-L SUM respectively. AlScore refers to AlignScore.}
\label{tab:internal_benchmarks}
\end{table*}

\begin{table*}[t!]
\centering
\scriptsize
\resizebox{\linewidth}{!}{%
\begin{tabular}{l|ccc|ccc|ccccc}
\toprule
\multicolumn{1}{c}{} & \multicolumn{3}{c}{\textbf{bitext\_custcomer\_support}} & \multicolumn{3}{c}{\textbf{banking77}} & \multicolumn{5}{c}{\textbf{DialogSum summarization}}  \\ 
\cmidrule(r){1-1}
\cmidrule(r){2-4} \cmidrule(r){5-7} \cmidrule(r){8-12}

\textbf{Models}  & \textbf{Precision} & \textbf{Recall} & \textbf{F1-micro} & \textbf{Precision} & \textbf{Recall} & \textbf{F1-micro} & \textbf{R1}  & \textbf{R2} & \textbf{RL} & \textbf{RLsum}  & \textbf{AlScore}   \\
\midrule
Qwen2-7b + \textbf{SFT-16bit}  & 0.975 & 0.975  & 0.975  & 0.8367  & 0.8367 & 0.8367 & 0.5249 & 0.2825 & 0.4312 & 0.4313 & 0.921 \\
Qwen2-7b + \textbf{PTQ-BNB-4bit}  & 0.975 & 0.975  & 0.975  & 0.8383  & 0.8383 & 0.8383 & 0.5264 & 0.2819 & 0.4303 & 0.4303 & 0.923 \\
Qwen2-7b + PTQ-BNB-4bit + \textbf{QLoRA}  & \textbf{0.995} & \textbf{0.995}  & \textbf{0.995}  & \textbf{0.905}  & \textbf{0.905}  & \textbf{0.905} & 0.5466 & 0.302  & 0.4523 & 0.4523 & \textbf{0.934} \\
Qwen2-7b + \textbf{PTQ-GPTQ-4bit}  & 0.9767  & 0.9767 & 0.9767 & 0.8417  & 0.8417 & 0.8417 & 0.522  & 0.2829 & 0.4289 & 0.4288 & 0.924 \\
Qwen2-7b + PTQ-GPTQ-4bit + \textbf{QLoRA} & \textbf{0.995} & \textbf{0.995}  & \textbf{0.995}  & \textbf{0.905}  & \textbf{0.905}  & \textbf{0.905} & \textbf{0.5474} & \textbf{0.3021} & \textbf{0.4533} & \textbf{0.4534} & 0.933 \\
\midrule
Llama2-7b + \textbf{SFT-16bit}  & 0.9967  & 0.9967 & 0.9967 & 0.8817 & 0.8817 & 0.8817 & \textbf{0.5816} & \textbf{0.3383} & \textbf{0.4875} & \textbf{0.4879} & \textbf{0.942} \\
Llama2-7b + \textbf{PTQ-BNB-4bit}  & 0.9967 & 0.9967 & 0.9967 & 0.8883  & 0.8883 & 0.8883 & 0.5739 & 0.3331 & 0.4813 & 0.4814 & 0.94 \\
Llama2-7b + PTQ-BNB-4bit + \textbf{QLoRA} & \textbf{0.9983}  & \textbf{0.9983} & \textbf{0.9983} & \textbf{0.9167}  & \textbf{0.9167} & \textbf{0.9167} & 0.5737 & 0.3293 & 0.4801 & 0.4803 & 0.938 \\
Llama2-7b + \textbf{PTQ-GPTQ-4bit}  & 0.9967  & 0.9967 & 0.9967 & 0.8817  & 0.8817 & 0.8817 & 0.5676 & 0.3226 & 0.4733 & 0.4736 & 0.934 \\
Llama2-7b + PTQ-GPTQ-4bit + \textbf{QLoRA}  & \textbf{0.9983} & \textbf{0.9983} & \textbf{0.9983} & 0.8983  & 0.8983 & 0.8983 & 0.5704 & 0.3266 & 0.4757 & 0.4757 & 0.938 \\
\midrule                                           
Mistral-7b + \textbf{SFT-16bit}  & 0.9983  & 0.9983 & 0.9983 & 0.9067  & 0.9067 & 0.9067 & 0.569  & 0.3331 & 0.4799 & 0.48 & 0.946 \\
Mistral-7b + \textbf{PTQ-BNB-4bit}  & 0.9983  & 0.9983 & 0.9983 & 0.905 & 0.905  & 0.905 & \textbf{0.5789} & \textbf{0.3394} & \textbf{0.487}  & \textbf{0.487} & \textbf{0.948} \\
Mistral-7b + PTQ-BNB-4bit + \textbf{QLoRA} & 0.9983    & 0.9983 & 0.9983 & 0.9033  & 0.9033 & 0.9033 & 0.5716 & 0.33   & 0.4786 & 0.4786 & 0.932 \\
Mistral-7b + \textbf{PTQ-GPTQ-4bit} & 0.9983  & 0.9983 & 0.9983 & 0.9033 & 0.9033 & 0.9033 & 0.5695 & 0.3312 & 0.4767 & 0.4766 & 0.947 \\
Mistral-7b + PTQ-GPTQ-4bit + \textbf{QLoRA}   & 0.9983  & 0.9983 & 0.9983 & \textbf{0.91} & \textbf{0.91} & \textbf{0.91} & 0.5678 & 0.3261 & 0.4749 & 0.4754 & 0.935 \\
\bottomrule
\end{tabular}
}

% \vspace{-1mm}
% \caption{\small{Performance of LLMs in zero-shot on the QMSUM dataset in the multi-query and single-query settings.}}
\caption{\small Performance of different models on the external task benchmark. R1, R2, RL and RLsum refer to ROUGE-1, ROUGE-2, ROUGE-L and ROUGE-L SUM respectively. AlScore refers to AlignScore.}
\label{tab:external_benchmarks}
\end{table*}

\subsubsection{Dataset Compilation}
To assemble the datasets for training and evaluation, both internal and external task datasets are combined with the General Instruction Dataset respectively. This is to ensure the model develops general instruction-following capability during both internal and external task fine-tuning processes. 

For evaluation purposes, as this study is focused on specific task performance, the General Instruction Dataset is thus excluded from the test split.
Table~\ref{tab:external_datasets_stats} presents detailed information on the sizes of all the datasets curated and used in our experiments.

\subsection{Training Hyperparameters and Setup}
For all three models and datasets, the maximum input context length is set to 3200 tokens and output to 800 tokens. Necessary filtering is applied to ensure our datasets fit with this context length limitation. Each fine-tuning job is conducted with two epochs on the dataset. Appendix \ref{sec:training-param} details other hyperparameters we apply for the fine-tuning process.

The fine-tuning and evaluation processes in our experiments are conducted using the Huggingface framework on a single node instance with 8 Nvidia A100 GPUs.

\subsection{Results}

Accuracy performance is evaluated at three different stages of the PTQ-QLoRA integration:
\begin{enumerate}
    \item 16-bit fully fined-tuned model after SFT, noted as \textbf{SFT-16bit}
    \item 4-bit quantized model on top of SFT, noted as \textbf{PTQ-\{quant-method\}-4bit}
    \item A LoRA with the 4-bit quantized model after the QLoRA fine-tuning, noted as \textbf{PTQ-\{quant-method\}-4bit+QLoRA}
\end{enumerate}
 We present our evaluation results on both internal and public datasets in Table~\ref{tab:internal_benchmarks} and Table~\ref{tab:external_benchmarks} respectively.  We perform Wilcoxon signed-rank test (p<=0.05) \cite{dror-etal-2018-hitchhikers} to compare whether the performance differences between PTQ-QLoRA and PTQ results for different models are statistically significant and find that they are significant for both classification (p=0.00047) and text generation tasks (p=0.004, 0.018, 0.034, 0.016 for ROUGE-1, -2, -L and -L SUM respectively). The performance difference between PTQ-QLoRA and 16-bit SFT is statistically significant for classification tasks (p=0.005) but not text generation tasks. The difference in performance between SFT and PTQ models is not statistically significant. In addition, we apply AlignScore \cite{zha-etal-2023-alignscore} on the summarization tasks to validate the factual consistency. The differences in factual consistency (based on AlignScore) are found not to be statistically significant. Further, we did not observe significant discrepancy between the models in format following or instruction following and therefore we omit the results of this evaluation. Based on this, our observations and findings can be summarized as follows:

\begin{enumerate}[(i)]
    \item The best accuracy performance is generally achieved by either the PTQ-QLoRA integration or the 16-bit full fine-tuning. This is consistent across all three base LLMs in our experiments. In other words, the PTQ-QLoRA integration can match and in many cases outperform 16-bit full fine-tuning in our target task performance.
    \item Applying quantization with or without additional QLoRa step does not significantly affect factual consistency on text generation tasks.
    \item In nearly all tasks, incorporating the QLoRA process enhances the accuracy of PTQ, regardless of the base model or the quantization method employed.
    \item Between the two quantization methods used in our experiments (BNB and GPTQ), we do not find a clear advantage of one method over the other. The relative performance difference can be affected by the base pre-trained model or the target task.

\end{enumerate}

\section{Conclusion}
In this study, we explore the PTQ-QLoRA that integrates 4-bit post-training quantization with QLoRA to optimize the deployment of LLMs in resource-limited environments. Through extensive experimentation, we demonstrate that this integration can match or surpass the performance of 16-bit full parameter fine-tuning, across various base LLMs, quantization methods and tasks. 

The results highlight that combining PTQ with QLoRA enhances model efficiency without sacrificing task-specific accuracy. This effective solution allows high-performing LLMs to be deployed with fewer resources. Overall, our findings underscore the potential of this integration to improve the practical deployment of LLMs, offering a scalable approach for future applications.

\section{Limitations}

A notable limitation of this work is that we do not compare the performance of applying QLoRA fine-tuning to a quantized base model prior to fine-tuning on the target dataset. In our limited experiments with this setting the resulting models consistently underperformed in comparison to both PTQ and PTQ-QLoRA, therefore we left this comparison out of the scope of this paper.

Further, we do not experiment with other bit precision levels and only use 4-bit quantization. Similarly to the above, our limited experiments have shown that currently 4-bit quantization is the most promising in terms of a trade-off between accuracy, inference performance, and available supporting infrastructure. In addition, we do not consider other quantization methods besides bitsandbytes and GPTQ for the reasons we explain in \ref{sec:quant_methods}. A more fine-grained look into different quantization methods and bit precision levels can be beneficial.

We also only experiment with several decoder-only models of the same size (7B) in this work as explained in \ref{sec:method_models} and are not considering the effects of quantization on the models with different architectures or number of parameters.

Finally, we benchmark the models on a limited number of tasks relevant to our business requirements and use autometrics for comparison. While we complement standard for text generation tasks ROUGE scores with a factual consistency metric AlignScore, a human review can reveal meaningful differences in performance between the models. Inclusion of other tasks as well as detailed evaluation of the outputs may be advantageous to understanding the benefits and limitations of our proposed technique.

\section{Ethical Considerations}

We maintained the licensing requirements accordingly while using open-source models and other tools from the providers
(e.g. OpenAI, Meta, Alibaba, Mistral, HuggingFace, etc.). Publicly available external datasets were used in our experiments only for evaluation and reproducibility purposes.

\bibliography{custom}

\appendix

\section{Appendix}
\label{sec:appendix}

\subsection{General Instruction Dataset}
\label{sec:instruct-data}
We adopt a similar approach as self-instruct \cite{wang-etal-2023-self-instruct} to generate instruction-following samples in the general domain. We start from manually creating 200 seed questions and generate ~50k instructions through bootstrapping as described in \cite{wang-etal-2023-self-instruct} using GPT-4. After necessary post-processing and filtering, GPT-4 is leveraged again to generate responses for each of the instructions. We provide some examples of the instructions in our General Instruction Dataset as follows:

\begin{itemize}
    \item Brainstorm a list of possible New Year's resolutions.
    \item Plan a weekly lunch menu for a school. Write down a main dish, a carbohydrate side dish, a vegetable side dish, and a dessert for each day.
    \item Translate the English sentence into Chinese: She went to school on Monday but found no other students, so she realized that Monday was actually a national holiday.
\end{itemize}

\subsection{Prompt Format for Internal Tasks}
\label{sec:fine-tuning-prompt}
The prompts we utilize for our internal tasks are as follows:

Summarization: \\
{\fontfamily{qcr}\selectfont{\normalsize \small Write a short and concise summary of the following conversation transcript focusing only on work or business-related topics without assessing its quality.\\
Transcript: \{\}
    }}\\

Note that we apply various summary length and style requirements in the prompt, such as long, medium, short, or bullet points.\\
\\

Action Items: \\
{\fontfamily{qcr}\selectfont{\normalsize \small You are provided with some text enclosed by curly brackets "\{\}", generate a newline-separated list of work, business or service-related TODO tasks that are still not done at the end of the conversation and should be completed after the conversation. Each task is a one-sentence summary of the action to be taken.\\
Transcript: \{\}
    }}\\

Call Purpose: \\
{\fontfamily{qcr}\selectfont{\normalsize \small For the conversation below, identify a single category for the purpose of the conversation chosen from this list: Account Management,
Appointment,
Billing Questions,
Callback,
Cancellation,
Claim,
Complaint.\\
Transcript: \{\}
    }}\\

Note that this is not the exhaustive list of the call purpose categories we support.\\

Call Outcome: \\
{\fontfamily{qcr}\selectfont{\normalsize \small For the conversation below, apply the appropriate category from the list provided below to describe the outcome of the conversation. Respond with "Other" if no category applies.: Call back,
Unsuccessful contact,
Voicemail Success,
Payment / Billing,
Status update,
Scheduled appointment,
Cancellation. \\
Transcript: \{\}
    }}\\
    
Note that this is not the exhaustive list of the call outcome categories we support.

\subsection{Training Hyperparameters}
\label{sec:training-param}
We provide the detailed hyperparameters we employ to fine-tune the LLMs in Table \ref{tab:hyperparameters}.

\begin{table}[h]
\centering
\scriptsize
\begin{tabular}{l|cc|cc}
\toprule
\multicolumn{1}{c}{} & \multicolumn{2}{c}{\textbf{Learning rate}} & \multicolumn{2}{c}{\textbf{Scheduler}}  \\ 
\cmidrule(r){1-1}
\cmidrule(r){2-3} \cmidrule(r){4-5}

\textbf{Models} & \textbf{Int} & \textbf{Ext} & \textbf{Int}  & \textbf{Ext} \\
\midrule
Qwen2-7B-SFT        & 3e-5      & 3e-5 & linear    & cosine   \\
\quad + BNB-4bit + QLoRA  & 3e-5      & 3e-5 & cosine    & cosine   \\
\quad + GPTQ-4bit + QLoRA & 3e-5      & 3e-5 & cosine    & cosine   \\
Llama2-7B-SFT       & 6e-6      & 6e-6 & linear    & linear   \\
\quad + BNB-4bit + QLoRA  & 2e-4      & 5e-4 & cosine    & linear   \\
\quad + GPTQ-4bit + QLoRA & 5e-4      & 5e-4 & cosine    & linear   \\
Mistral-7B-v0.3-SFT & 6e-6      & 6e-6 & linear    & linear   \\
\quad + BNB-4bit + QLoRA  & 5e-4      & 5e-4 & linear    & linear   \\
\quad + GPTQ-4bit + QLoRA & 5e-4      & 5e-4 & linear    & linear  \\
\bottomrule
\end{tabular}
\caption{\small Training hyperparameters for internal (Int) and external (Ext) datasets.}
\label{tab:hyperparameters}
\end{table}

\end{document}